\documentclass[letterpaper, 10 pt, journal, twoside]{IEEEtran}
\usepackage{amsmath,amsfonts}
\usepackage{algorithmic}
\usepackage{array}
\usepackage[caption=false,font=normalsize,labelfont=sf,textfont=sf]{subfig}
\usepackage{textcomp}
\usepackage{stfloats}
\usepackage{url}
\usepackage{verbatim}
\usepackage{graphicx}

\usepackage[table,xcdraw]{xcolor}
\usepackage{graphicx}
\usepackage{tabularx}
\usepackage{amsmath}
\usepackage{mathtools}
\usepackage{mathtools}
\usepackage{multirow}
\usepackage{hhline}
\usepackage{booktabs}
\usepackage[]{algorithm2e}
\usepackage{mathabx}
\usepackage{array}
\usepackage{comment}
\usepackage{amssymb} % define this before the line numbering.
\usepackage{color}
\usepackage{cite}
\usepackage{url}

\def\eg{\emph{e.g}.}

\newcommand{\Tref}[1]{Table~\ref{#1}}
\newcommand{\Eref}[1]{(\ref{#1})}
\newcommand{\Fref}[1]{Fig.~\ref{#1}}

\begin{document}
\title{
ScaTE: A Scalable Framework for Self-Supervised Traversability Estimation in Unstructured Environments
}

% \author{Junwon Seo$^*$, Taekyung Kim$^*$, Kiho Kwak, Jihong Min, and Inwook Shim$^\dagger$
% \thanks{Junwon Seo, Taekyung Kim, Kiho Kwak, and Jihong Min are with the Agency for Defense Development, Daejeon 34186, Republic of Korea (e-mail: junwon.vision@gmail.com, tkkim.robot@gmail.com, kkwak.add@gmail.com, happymin77@gmail.com).}%
% \thanks{Inwook Shim is with Department of Smart Mobility Engineering, Inha University, Republic of Korea (e-mail: iwshim@inha.ac.kr).}%
% \thanks{$^{*}$These authors contributed equally to this work.}%
% \thanks{$^{\dagger}$Corresponding author.}%
% \thanks{The multimedia material is available at https://youtu.be/kSvqjHDqmIk.}
% }

%September 14, 2022 
%November 14, 2022
%December 12, 2022

\author{Junwon Seo, Taekyung Kim, Kiho Kwak, Jihong Min, and Inwook Shim%

\thanks{Manuscript received: September, 14, 2022; Revised November, 14, 2022; Accepted December, 12, 2022. This paper was recommended for publication by Editor E. Marchand upon evaluation of the Associate Editor and Reviewers' comments. This work was supported by the Agency For Defense Development, funded by the Korean Government in 2022 and Korea Institute for Advancement of Technologe~(KIAT) grant funded by the Korea Government~(MOTIE) (P0020536, HRD Program for Industrial Innovation).  \textit{(Junwon Seo and
Taekyung Kim are co-first authors.)} {(Corresponding author: Inwook Shim.)}}
\thanks{Junwon Seo, Taekyung Kim, Kiho Kwak, and Jihong Min are with the Agency for Defense Development, Daejeon 34186, Republic of Korea
        (email:junwon.vision@gmail.com, tkkim.robot@gmail.com, kkwak.add@gmail.com, happymin77@gmail.com).}%
\thanks{Inwook Shim is with the Department of Smart Mobility Engineering, Inha University, Republic of Korea
        (email:iwshim@inha.ac.kr)}%
\thanks{The multimedia material is available at https://youtu.be/ZZHfD-8OpBg.}
\thanks{Digital Object Identifier (DOI): 10.1109/LRA.2023.3234768}
}

\markboth{IEEE Robotics and Automation Letters. Preprint Version. ACCEPTED DECEMBER, 2022}
{Seo \MakeLowercase{\textit{et al.}}: ScaTE: A Scalable Framework for Self-Supervised Traversability Estimation in Unstructured Environments} 
\maketitle
% \thispagestyle{empty}
% \pagestyle{empty}

%%%%%%%%%%%%%%%%%%%%%%%%%%%%%%%%%%%%%%%%%%%%%%%%%%%%%%%%%%%%%%%%%%%%%%%%%%%%%%%%
\begin{abstract}

For the safe and successful navigation of autonomous vehicles in unstructured environments, the traversability of terrain should vary based on the driving capabilities of the vehicles. Actual driving experience can be utilized in a self-supervised fashion to learn vehicle-specific traversability. However, existing methods for learning self-supervised traversability are not highly scalable for learning the traversability of various vehicles. In this work, we introduce a scalable framework for learning self-supervised traversability, which can learn the traversability directly from vehicle-terrain interaction without any human supervision. We train a neural network that predicts the proprioceptive experience that a vehicle would undergo from 3D point clouds. Using a novel PU learning method, the network simultaneously identifies non-traversable regions where estimations can be overconfident. With driving data of various vehicles gathered from simulation and the real world, we show that our framework is capable of learning the self-supervised traversability of various vehicles. By integrating our framework with a model predictive controller, we demonstrate that estimated traversability results in effective navigation that enables distinct maneuvers based on the driving characteristics of the vehicles. In addition, experimental results validate the ability of our method to identify and avoid non-traversable regions.

\end{abstract}


%%%%%%%%%%%%%%%%%%%%%%%%%%%%%%%%%%%%%%%%%%%%%%%%%%%%%%%%%%%%%%%%%%%%%%%%%%%%%%%%
\vspace{-1.3mm}
\section{Introduction}
\begin{figure}[t]
\centering
\includegraphics[width=0.99\linewidth]{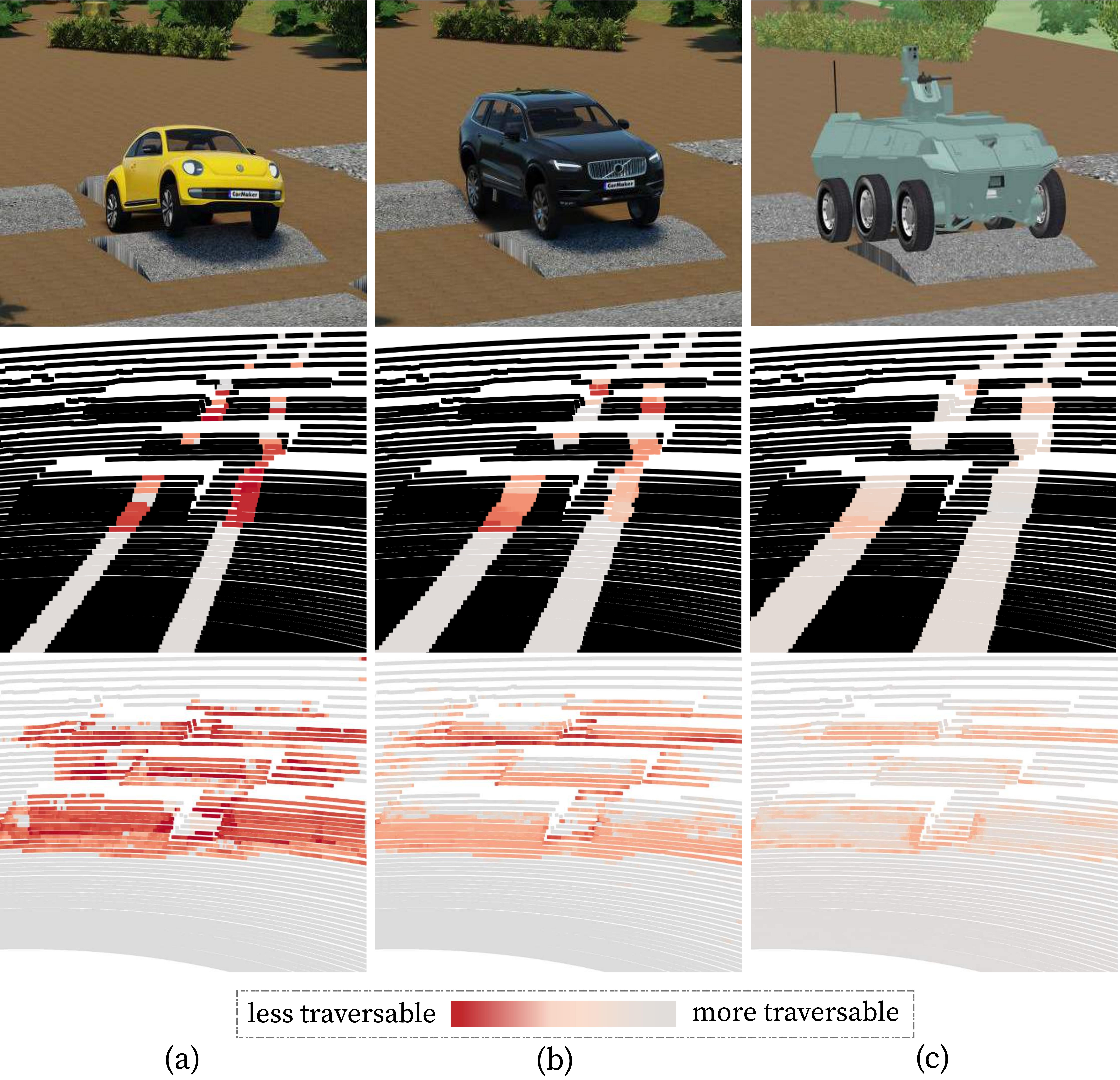}
\caption{{\textit{First} row: the appearance of the vehicles}, \textit{Second} row: self-supervised traversability data, \textit{Third} row: traversability estimation result for each vehicle. Vehicles with varying driving capabilities interact differently with the same terrain: (a) a compact car, (b) an SUV, (c) a $6\times6$ off-road vehicle.}
\label{fig:Main}
\vspace{-0.2in}
\end{figure}

\IEEEPARstart{E}{stimating} the traversability of a driving surface is an essential component of autonomous driving systems. Numerous studies in computer vision have made significant advances in identifying drivable surfaces from visual sensors. They are mostly operated in structured environments, such as urban and indoor scenes, where the traversability can be explicitly specified under human supervision. Using human-annotated and large-scale datasets~\cite{geiger2013vision}, the classification-based methods show remarkable results in learning the terrain traversability. However, for safe and reliable navigation in unstructured environments with a variety of challenging terrain, the estimated traversability should vary based on the driving capability of each vehicle, which depends on numerous vehicle characteristics~(\eg{, powertrain, vehicle weight, type and size of wheels, and shock absorbers}). Therefore, when estimating traversability, it is necessary to take into account not only the terrain characteristics but also the driving capacity of a vehicle, as shown in \Fref{fig:Main}, which is referred to as \textit{self-supervised} traversability estimation~\cite{kim2006traversability}.

Various works present the self-supervised approaches to estimating traversability utilizing the actual driving experiences instead of human-provided annotations~\cite{Dahlkamp-RSS-06,wellhausen2019should,sathyamoorthy2022terrapn,kolvenbach2019haptic,TERP,vibration,acoustic,kahn2021badgr,frey2022locomotion}. While these methods successfully estimate the traversability of the environment tailored for each vehicle, the methodologies are not \textit{scalable} for the following reasons. First, they still involve a degree of human supervision, such as manual labeling and an expert-designed definition of traversability. Whenever annotation classes, the type of a vehicle, or the sensor type of a vehicle are changed, the manual data annotation process must be re-performed. In addition, the inflexible definition of traversability may not be suitable for all kinds of vehicles and operational objectives. Second, they do not clearly address the uncertainty problem inherent in self-supervised traversability estimation. Due to the fact that self-supervised data only provides supervision for actually traversed regions, estimates for non-traversed regions are subject to high uncertainty, which can lead to unreliable navigation~\cite{schmid2022self}. Lastly, they do not conduct experiments to verify the scalability of their methods with various vehicles.

Scalability is the key issue when it comes to learning self-supervised traversability for various vehicles. To achieve a high level of scalability, the network should be self-trained without any external human annotation. Otherwise, the annotation process would be too laborious, or the human supervision could erroneously represent a vehicle's driving capability. Additionally, vehicle-terrain interaction should be directly estimated instead of a handcrafted traversability cost. This can lead to precise and reliable navigation in unstructured environments for various vehicle types and navigational policies.

\begin{figure}[t]
\centering
\includegraphics[width=0.99\linewidth]{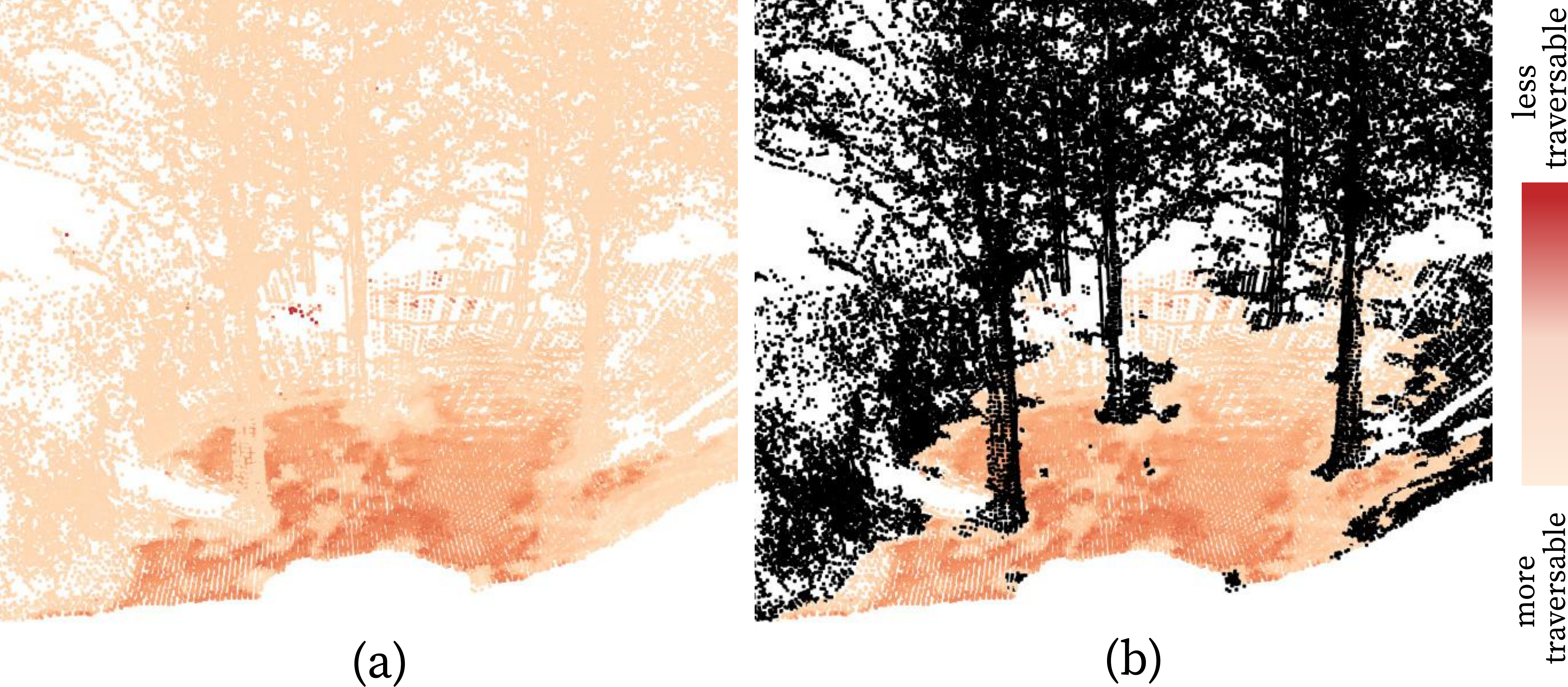}
\caption{Visualization of the uncertainty problem of the self-supervised traversability estimation. (a)~Estimations can be overconfident on non-traversed regions. Bushes, gravels, steep slopes, and trees are estimated as highly traversable. (b)~The regions should be identified explicitly as non-traversable so that they can be avoided during navigation.}
\label{fig:Uncertainty}
\vspace*{-0.3in}
\end{figure}

In this paper, we propose a scalable framework for learning self-supervised traversability. Since the framework is trained without human supervision, it is capable of learning the traversability of various vehicles.Our framework predicts experiences that a vehicle would undergo while traversing the terrain. It can provide precise and flexible information on where the vehicle can traverse and how challenging it would be for safe and effective navigation. Moreover, with a novel positive-unlabeled~(PU) learning method~\cite{bekker2020learning} that does not require any aid from humans, our framework addresses the uncertainty problem that arises when self-supervised traversability data is used. For reliable estimation, our framework identifies regions with high epistemic uncertainty, guiding the navigation policy to avoid such terrain.

To demonstrate the efficacy of our framework, we conduct experiments with vehicles of varied driving capabilities using a high-fidelity vehicle simulator. The traversability estimation results enable vehicles with differing driving abilities to take differentiated maneuvers, resulting in effective navigation in unstructured environments. In addition, we validate our method using real-world driving data collected in challenging unstructured environments.

\section{Related Works}
\subsection{Traversability Estimation with Human Supervisions}
Advances in learning traversability have resulted in significant progress in autonomous navigation in various environments. Early works for estimating traversability primarily focus on analyzing simple geometric features of environments such as slopes, steps, and roughness using elevation maps~\cite{navigation2016, fankhauser2014robot} and the surface normal of the terrain obtained from the local distribution of point clouds~\cite{ahtiainen2017normal}. With the advent of deep neural networks, semantic segmentation is used to classify terrain into predetermined terrain classes or traversable and non-traversable regions using a large-scale dataset~\cite{jiang2021rellis}. In unstructured environments, such as off-road, various methods have been developed to estimate the static terrain traversability~\cite{sock2016probabilistic, krusi2017driving,semanticOffroad,TNS,Fan2021STEPST}. However, such methods rely on human supervision about terrain rather than a vehicle's actual driving capability, necessitating laborious and potentially erroneous manual annotation procedures. The vehicle-agnostic traversability is unlikely to function successfully in unstructured environments.
\subsection{Self-Supervised Traversability Estimation}
For reliable navigation in unstructured environments, traversability should be jointly determined by a vehicle and terrain~\cite{kim2006traversability}. Recent research has focused on self-supervised learning approaches that exploit a vehicle's driving experience to learn the traversability of a terrain. Sensor readings from onboard proprioceptive sensors, such as force-torque sensors~\cite{wellhausen2019should,kolvenbach2019haptic}, Inertial Measurement Unit~(IMU)~\cite{sathyamoorthy2022terrapn,kahn2021badgr}, and acoustic sensors~\cite{vibration} are utilized to capture the interactions between vehicles and terrain. Additionally, the random traversal results from simulations~\cite{frey2022locomotion}, odometry errors~\cite{sathyamoorthy2022terrapn}, and attention mask~\cite{TERP} are used to model the vehicle-terrain interaction.

However, these methods either require manual labeling to classify terrain into predetermined classes~\cite{wellhausen2019should,vibration}, depend on hand-crafted heuristics to assess navigational cost~\cite{sathyamoorthy2022terrapn, kolvenbach2019haptic}, or rely on simulation for training, which is only applicable to a specific type of vehicle and sensor configuration. Moreover, the vehicle-terrain interaction data are confined to traversed regions only, and gathering data from non-traversable areas is infeasible~\cite{Wellhausen_2020}. This characteristic induces epistemic uncertainty in the estimation, which hinders the reliability of navigation~\cite{schmid2022self}. For example, bumps, bushes, and gravel with diverse unknown shapes that a vehicle has never experienced can be overestimated as highly traversable, resulting in catastrophic navigational failure.
\subsection{One-Class Classification}
Anomaly detection is utilized to identify regions or circumstances with a high degree of uncertainty in autonomous navigation~\cite{Wellhausen_2020,pmlr-v155-ji21a}. Deep learning-based anomaly detection methods only incorporate normal data for learning a one-class classifier, making them prone to a hypersphere collapse solution in which the network is trivially taught to classify all inputs as normal~\cite{ruff2018deep}. Positive unlabeled learning learns a binary classifier from data containing only positive and unlabeled examples~\cite{bekker2020learning}. The risk-estimator-based algorithms assume unlabeled data are a mixture of positive and negative data and minimize the empirical risk of classification, which is indirectly calculated from unlabeled data~\cite{kiryo2017positive}. Their performance is heavily dependent on the assumptions regarding the distribution of data, which are usually unknown in practice~\cite{bekker2020learning}. The unlabeled data can be used in an unsupervised manner to learn discriminative feature representations~\cite{caron2020unsupervised,asano2020self}. Since the self-supervised traversability data consist of positive and unlabeled data, our method addresses the uncertainty problem using a novel PU learning method without external annotation.

\section{ScaTE: Scalable Traversability Estimation}

% \subsection{Overview}
We propose a scalable framework for learning self-supervised traversability. First, the self-supervised traversability data of PU type are generated by an automated procedure. Note that the data consist solely of raw sensor measurements from onboard sensors, with no human annotations or human-designed definition of traversability. Then, the network is trained to predict traversability from 3D point clouds. Learning traversability with just PU-type data leads to uncertainty in estimation, as illustrated in \Fref{fig:Uncertainty}. Therefore, regions with high uncertainty are identified to be avoided during navigation.
\subsection{Automated Data Generation}
We utilize point clouds to capture the geometric properties of the environments. Vehicle-terrain interaction data are gathered using proprioceptive sensors, and the collected sensor signals are automatically mapped to the point cloud without human intervention. To precisely model the interaction, the experience of each wheel is recorded separately. The vehicle-terrain interaction for each wheel can supervise effective navigation and be employed for various navigation strategies. For instance, the vertical force exerted on the wheels can offer precise information regarding the difficulty and stability of traversing~\cite{tire}.

The vehicle's interaction with the terrain is projected onto point clouds by locating the wheel contact points. For precise mapping, LiDAR-based SLAM is used to recover poses and wheel trajectories. Wheel contact regions are computed from the poses, and points closer to the contact regions than the wheel thickness are considered traversable and labeled as \textit{positive} with the ground-truth traversability, while the remaining points are left \textit{unlabeled}. Additionally, the state of the vehicle at the time of the contact is recorded. We include speed and heading angle since they can illuminate how the vehicle interacts with the terrain during movement.

In summary, the self-supervised data of the PU type, $\mathbf{P}=\{\mathbf{P}_p, \mathbf{P}_u\}$, consist of a relatively small number of positive points, $\|\mathbf{P}_p \| = n_p$, and unlabeled points, $\|\mathbf{P}_u \| = n_u$.
\subsection{Learning Traversability}
\begin{figure}[t]
\centering
\includegraphics[width=0.99\linewidth]{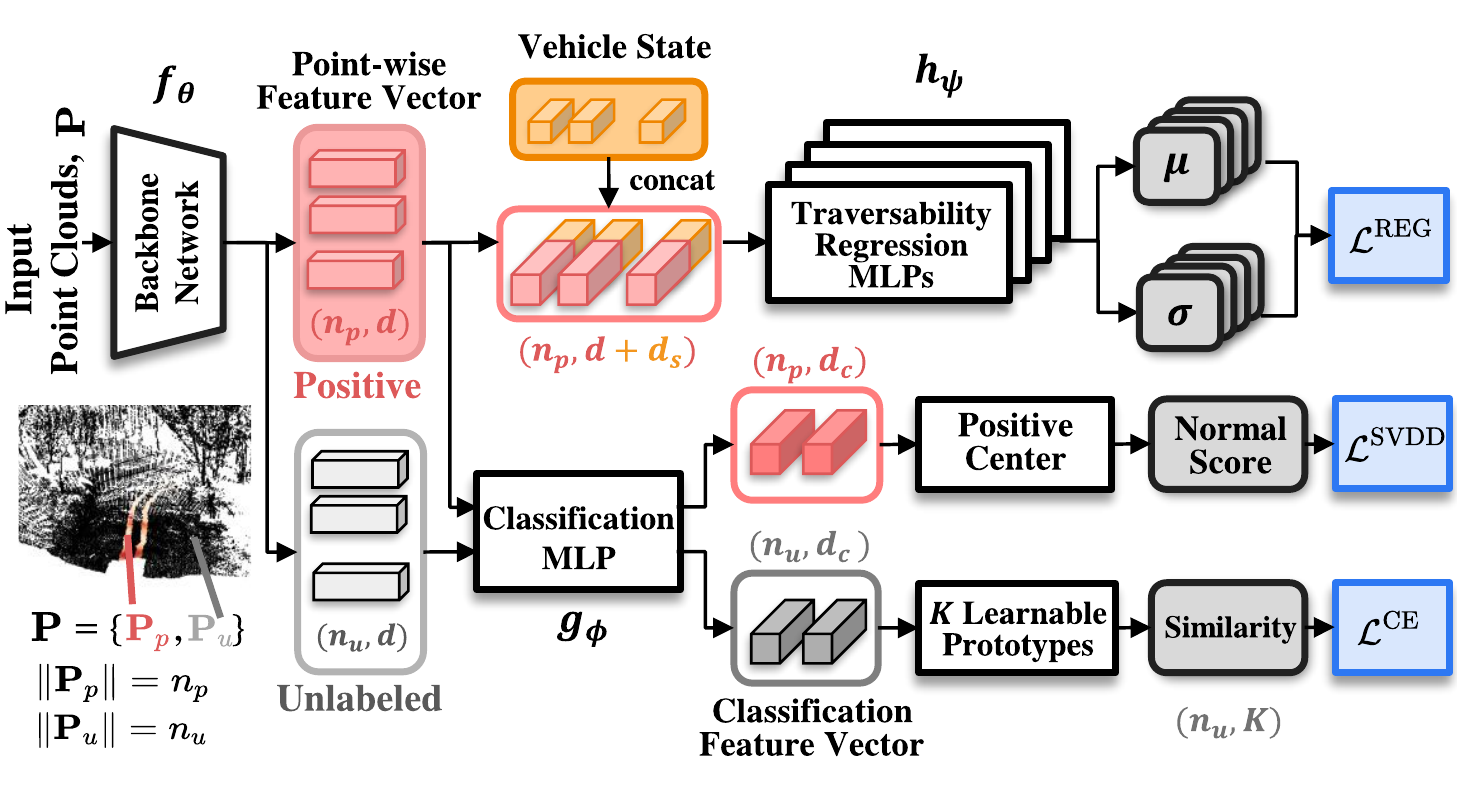}
\caption{High-level structure of the proposed learning framework.}
\label{fig:architecture}
\vspace*{-0.2in}
\end{figure}

Given the representation of environments as 3D point clouds, we aim to learn a model that simultaneously estimates point-wise regression of \textit{traversability} and \textit{binary classification} that identifies whether the terrain is traversable or not. Specifically, we aim to find a mapping ${(\mathbf{T}, \mathbf{Y}) = \mathcal{F} (\mathbf{P}, \mathbf{S})}$, where ${\mathbf{P} \in \mathbb{R}^{n \times 3}}$ denotes $3$D points, $\mathbf{S}$ indicates the states of a vehicle when interacting with the contact points, and $n = n_p + n_u$ indicates total number of points. Traversability $\mathbf{T}$ represents the interaction between the wheels of a vehicle and the terrain when the vehicle traverses the terrain. Here, the interaction $\mathbf{T}$ can be defined in various ways using proprioceptive sensor measurements. The binary mask $\mathbf{Y}\in\{0,1\}^{n}$ specifies whether or not the estimations on the points are reliable, namely traversable. When performing navigation tasks, the points estimated as non-traversable are avoided to prevent undesirable maneuvers.
\begin{align}
    \mathbf{x}_i &= f_\theta(\mathbf{p}_i), \label{eq1}\\
    \mathbf{T}_i &= h_\psi(\mathbf{x}_i, \mathbf{s}_i), \mathbf{Y}_i = g_\phi(\mathbf{x}_i) \label{eq3}.
\end{align}
The framework is trained with separate heads that take the point-wise embedding vector from the point-cloud encoding backbone network as input and output the traversability regression vector and the binary classification mask. Let $f_\theta$ denotes the $3$D point cloud encoder that maps a point $\mathbf{p}_i\in\mathbb{R}^{3}$ to point-wise feature $\mathbf{x}_i \in \mathbb{R}^{d}$, where $\theta$ is a network parameter. RandLA-Net~\cite{hu2020randla} is used for the backbone network due to its ability to efficiently compute per-point features for large-scale point clouds in real-time. $h_\psi$ and $g_\phi$ are multi-layer perceptron~(MLP) heads for traversability regression and binary classification, respectively, that share the backbone encoder. Each MLP head consists of two layers with a dropout in the middle, and the traversability regression head exists independently for each wheel. Note that we use both positive and unlabeled data for classification but only positive data for traversability regression. The overall architecture of the training procedure of our learning framework is illustrated in \Fref{fig:architecture}.

With positive data, point feature $\mathbf{x}_i$ extracted from the backbone are concatenated with vehicle state, $\mathbf{s}_i\in\mathbb{R}^{d_S}$. The concatenation is conducted after point feature encoding so that model predictive controllers~\cite{SMPPI}, which sample numerous trajectories with various states, can efficiently utilize the estimation. Due to the stochastic nature of physical interaction, the regression MLP heads learn the mean and standard deviation of the traversability, $(\boldsymbol{\mu}_i, \boldsymbol{\sigma}_i) = h_\psi(\mathbf{x}_i, \mathbf{s}_i)$, and the network is trained to minimize the gaussian log-likelihood loss:
\begin{equation}\label{gaussiannll}
    \mathcal{L}^{\text{REG}}(\mathbf{x}_i) = \frac{1}{2}\left(\log(\boldsymbol{\sigma}_i) +  \frac{(\boldsymbol{\mu}_i - \mathbf{T}_i)^2}{\boldsymbol{\sigma}_i} \right).
\end{equation}

\subsection{Binary Classification for Uncertainty Handling}
The concept of anomaly detection is used for binary classification. The point-wise feature of positive data $\mathbf{x}_i$ is forwarded to the binary classification MLP head to generate classification feature vector, $g_\phi: \mathbb{R}^{d} \rightarrow \mathbb{R}^{d_C}$, which is utilized to classify whether or not the point is traversable. Similar to the Deep SVDD~\cite{ruff2018deep}, the classification feature embedding space is trained to minimize the volume of a positive-data-enclosing hypersphere centered on the positive-center, $\mathbf{C}_p \in \mathbb{R}^{d_C}$, using positive data only with \Eref{eq:svdd}. The normal score of data is derived from the similarity between the embedded classification feature and the positive center, and each point is classified as either traversable or non-traversable with a simple threshold:
\begin{equation}\label{eq:svdd}
    {\mathcal{L}^{\text{SVDD}}(\mathbf{x}_i) = \| g_\phi(\mathbf{x}_i) - \mathbf{C}_p \|^2 }.
\end{equation}

However, the solution is susceptible to a hypersphere collapse solution. With sole guidance of $\mathcal{L}^{\text{SVDD}}$ with positive data, the majority of the data can be trivially mapped to the positive center. In addition, the network does not make use of unlabeled data, which is not labeled as positive but observed by a sensor, and therefore cannot embed the entire data distribution effectively.

Inspired by unsupervised clustering~\cite{asano2020self}, our method exploits the unlabeled data in an unsupervised manner to make classification features for one-class classification more discriminative. The unlabeled data are clustered by mapping the point feature representations of unlabeled data, $\mathbf{x}_j$, to a set of $K$ learnable prototypes $\mathbf{C}_u = \{\mathbf{c}_1, \cdots, \mathbf{c}_K\}$, or cluster centers, by the following procedures. 

First, the posterior distribution of the unlabeled points to the prototypes, $\mathbf{Q} \in \mathbb{R}^{K \times n_u}$ is computed by taking the softmax of the similarity between features and prototypes, as \Eref{posterior} where $\tau$ is a temperature parameter and set to $0.05$:
\begin{equation}\label{posterior}
    \mathbf{Q}_{kj} = \frac{\exp(\mathbf{x}_j^\intercal \mathbf{c}_k / \tau)}{\sum_{k'=1}^{K}\exp(\mathbf{x}_j^\intercal \mathbf{c}_{k'} / \tau)}.
\end{equation}

Second, each point is assigned to one of the prototypes, and we denote the assignment of points to prototypes as ${\mathbf{A} \coloneqq [\mathbf{A}_j]_{j=1}^{n_u} \in \{0,1\}^{K \times n_u}}$, where $\mathbf{A}_{j} \in \{0, 1\}^K$ is the one-hot cluster assignment. To avoid trivial solutions in which all samples are mapped to a single prototype, an equipartition constraint is added to enforce the assignments to split points into subsets of equal size. The assignment matrix is obtained by maximizing the similarity between features and prototypes under the equipartition constraint:
\begin{equation}\label{assignment}
    \max_{\mathbf{A}} \operatorname{Tr}(\mathbf{A}^\intercal \mathbf{Q}) \quad s.t. \quad \mathbf{A}\cdot\mathbf{1}^{n_u} = \frac{n_u}{K}\cdot\mathbf{1}^{K},
\end{equation}
where $\mathbf{1}^{n_u}\in \mathbb{R}^{n_u}$ and $\mathbf{1}^{K}\in \mathbb{R}^{K}$ denotes the vector of all ones. This is an optimal transport problem and can be solved efficiently by a few iterations of the \textit{Sinkhorn-Knopp} algorithm~\cite{sinkhorn}, which outputs the re-normalization vectors $\mathbf{u} \in \mathbb{R}^{K}$ and  $\mathbf{v} \in \mathbb{R}^{n_u}$:
\begin{align}
& \mathbf{A} = \operatorname{Diag}(\mathbf{u})\exp\left(\frac{\mathbf{Q}}{\tau}\right)\operatorname{Diag}(\mathbf{v}).
\end{align}

Then, the features and prototypes are jointly updated to minimize the cross-entropy loss between the posterior distribution of points and the cluster assignment as follows:
\begin{equation}\label{CE}
    \mathcal{L}^{\text{CE}} = - \frac{1}{n_u} \sum_k^{K}\sum_j^{n_u} \mathbf{A}_{kj} \log (\mathbf{Q}_{kj}).
\end{equation}

By clustering the unlabeled data, data semantically similar to the positive data are embedded close to the positive center, while data semantically dissimilar from the positive data are embedded distantly from the center. As a result of the training, the embedding space not only avoids a trivial solution but also effectively reflects the whole data distribution. The final objective of our learning framework is as follows: 
\begin{equation}\label{OBJECTIVE}
    \mathcal{L} = \frac{1}{n_p}\sum_{\mathbf{P}_p}{(\mathcal{L}^{\text{REG}} + \mathcal{L}^{\text{SVDD}}}) + \frac{1}{n_u}\sum_{\mathbf{P}_u}{\mathcal{L}^{\text{CE}}}.
\end{equation}

\section{Experiments on a Vehicle Simulator}
In this section, we design an experiment to demonstrate the scalability of our framework using a high-fidelity simulator. It is challenging to collect driving data for multiple vehicles in identical circumstances, and it would be dangerous to collect various interaction data with a real vehicle over unstructured terrain. Therefore, the high-fidelity vehicle simulator, CarMaker, is utilized for the experiments. We qualitatively verify that our framework enables vehicles with different driving capabilities to learn distinctive traversability, resulting in effective navigation in unstructured environments. Additionally, we undertake an ablation study to show the significance of each component of our framework.
\begin{figure}[t]
\centering
\includegraphics[width=0.99\linewidth]{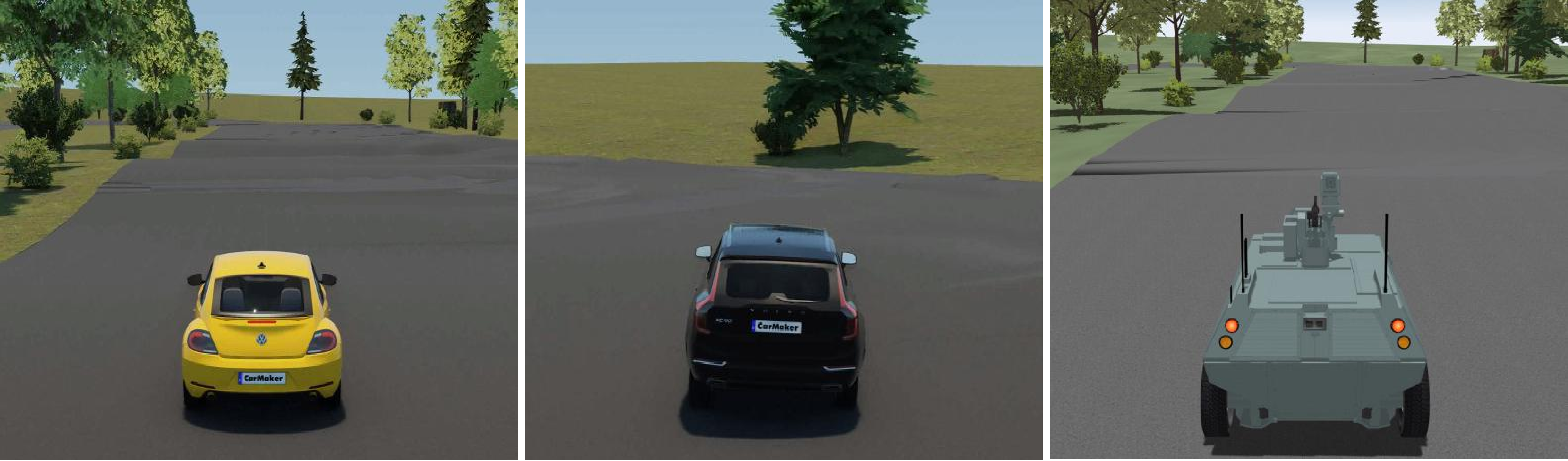}
\caption{The off-road scene used for training data collection. For three vehicles, interaction data are obtained and mapped to contact points of point clouds in a self-supervised manner. In order to obtain a variety of interaction data, vehicles are simulated to traverse through all terrains.}
\label{fig:training}
\vspace*{-0.2in}
\end{figure}

\begin{figure*}[t]
\centering
\includegraphics[width=0.99\linewidth]{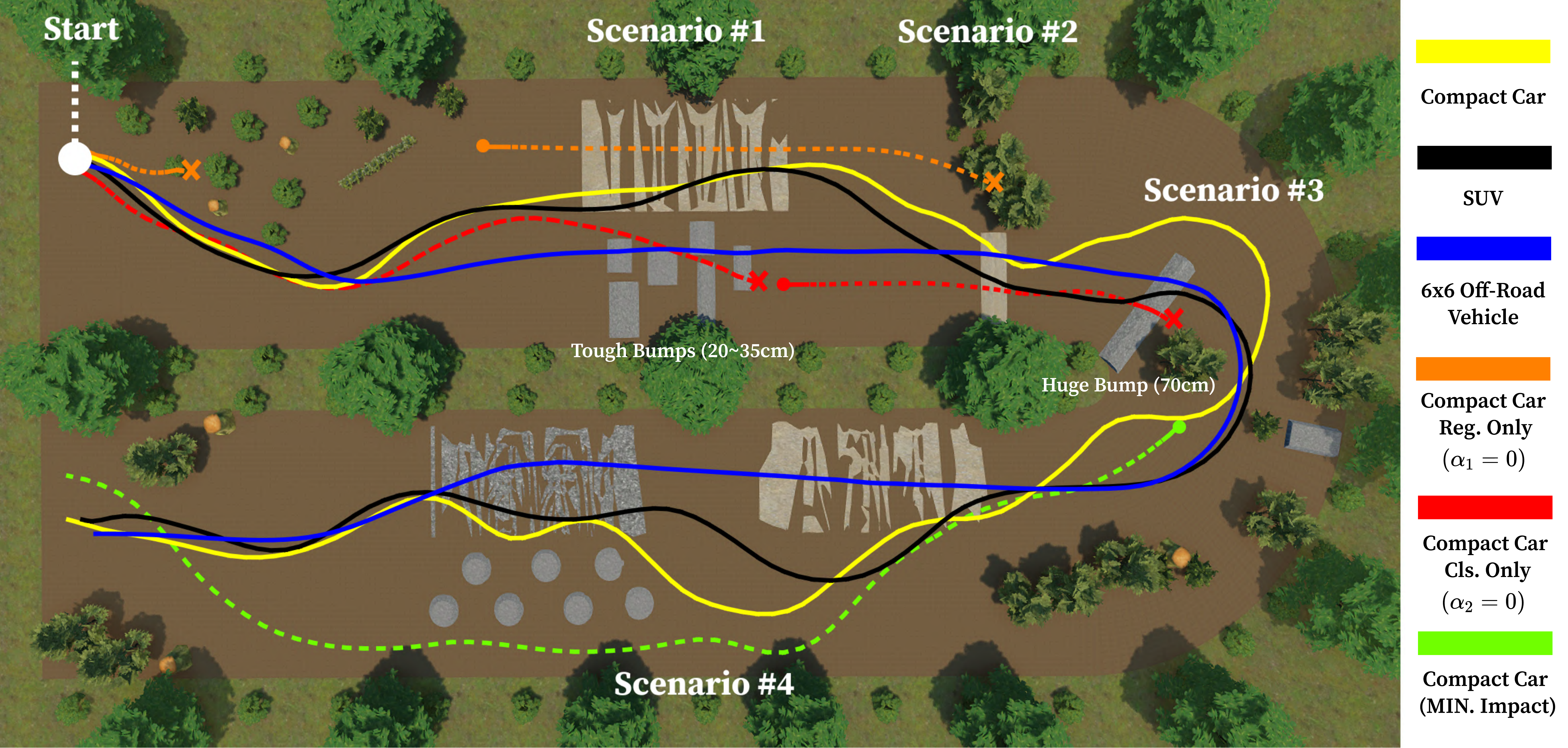}
\caption{Visualization of navigation results on the vehicle simulator. Colored lines represent trajectories, and dotted lines illustrate the results of the ablations.}
\label{fig:carmaker}
\vspace*{-0.2in}
\end{figure*}

\subsection{Experimental Setup}

An off-road scene with various types of unstructured components is designed for obtaining training data. Randomly patterned rough terrain, large bumps in various shapes, and potholes are generated, all of which vary in size and exist on both straight and curved roads. Additionally, non-traversable components such as trees, bushes, and wooden pillars are added, and such regions are never traversed by the vehicles. The driving data are obtained for three types of vehicles: a compact car, an SUV, and a $6\times6$ off-road vehicle, as shown in \Fref{fig:training}. In this experiment, we employ the vertical forces exerted on each wheel as ground-truth traversability. The $128$-layer LiDAR is utilized to acquire point clouds, and random rotation along the z-axis and random scaling are applied in terms of data augmentation.

For evaluation, we design a realistic off-road scene with unstructured components different from the training scene. The predicted vertical force on wheels and binary classification results on point clouds are transformed into a $2.5$D grid map. The sampling-based model predictive controller is used for navigation, utilizing the estimated traversability. We employ Smooth Model Predictive Path Integral (SMPPI) controller \cite{SMPPI} using the vehicles' dynamics models. Based on our prior work \cite{SMPPI}, we formulate a simple cost function $c(\mathbf{T}_i, \mathbf{Y}_i)$, which consists of two components as follows:
\begin{equation} \label{eq:cost}
    c(\mathbf{T}_i, \mathbf{Y}_i) = \alpha_1{\text{Uncertainty}(\mathbf{Y}_i)} + \alpha_2{\text{Stabilizing}(\mathbf{T}_i)}.
\end{equation}

The uncertainty cost imposes a significant penalty if the wheel-terrain contact point is non-traversable. This hard cost prevents the vehicle from traversing regions with high epistemic uncertainty. The stabilizing cost is an example of a cost function that makes use of the predictive vehicle experience. The roll and pitch impacts are derived from the combination of predicted vertical forces on wheels, and rotational motions of the vehicles are derived based on the vehicle's specifications. The sampled trajectories that are expected to undergo large rotational movements are assigned a large cost, hence promoting stable and secure driving across uneven yet still traversable terrain.

After learning the traversability of each vehicle, we perform navigation on the off-road scene for evaluation  with $\alpha_1=\alpha_2=0.5$. Using the estimated traversability of each vehicle, each vehicle takes different movements in the same scene for successful navigation. For fair comparisons, we maintain the identical parameters for the controller, the only difference being the predicted traversability. Furthermore, for ablation, navigation is undertaken in the absence of one of our components, traversability regression and binary classification, to validate the efficacy of each component. We employ the compact car for ablation, which demands the most precise control for stable driving through unstructured terrain. By setting the coefficient $\alpha_1=0$ or $\alpha_2=0$, the compact car is navigated without classification or regression, respectively. The results are shown in \Fref{fig:carmaker}, and detailed navigation results are available in the multimedia material.

\subsection{Experimental Results}
\subsubsection{Distinct Traversability for Vehicles}\label{sssec:num1}
The three vehicles navigate along different paths based on the estimated traversability. In \textit{Scenario~$\#1$}, the $6\times6$ off-road vehicle~(colored in \textit{blue}) traverses through the tough bumps that the compact car~(colored in \textit{yellow}) and even the SUV~(colored in \textit{black}) decide to circumvent. This is because the $6\times6$ off-road vehicle is capable of sustaining greater impact than other vehicles. The compact car takes sharp steering to avoid the huge bump on the curved road~(see \textit{Scenario~$\#3$}), which other vehicles are able to drive through. It verifies that our framework can estimate traversability for various vehicle types, resulting in differentiated navigational maneuvers.

\subsubsection{Without Binary Classification}\label{sssec:num2}
At the \textit{Start} point, vehicles encounter a group of non-traversable components such as bushes. Without the binary classification, they are estimated with overconfidence as being easily traversable. While vehicles with the binary classification deviate from such regions, the compact car without the uncertainty cost~(colored in \textit{orange}) fails to avoid the regions. In \textit{Scenario~$\#2$}, the non-traversable components are adjacent to a traversable bump. Without uncertainty handling, the non-traversable components are estimated as being more traversable than the bump, resulting in a collision. The binary classification clearly determines the shrubs as non-traversable, forcing the vehicle to make a detour~(colored in \textit{yellow}).

\subsubsection{Without Traversability Regression}\label{sssec:num3}
Without the traversability regression, the cost function just simply distinguishes whether the terrain is traversable or not, disregarding the terrain's precise level of navigational difficulty. In \textit{Scenario~$\#1$}, the compact car controlled without traversability regression~(colored in \textit{red}) follows a straight path to a series of tough bumps and experiences strong impacts, resulting in unsteady navigation and eventual vehicle overturning. Despite the fact that the tough bumps are still traversable, the vehicle with the stabilizing cost deliberately avoids the tough bumps. 
Additionally, as depicted in \textit{Scenario~$\#3$}, the stabilizing cost instructs the controller to take a detour when safer terrain exists in order to circumvent the huge bump~(colored in \textit{yellow}). However, the controller without the stabilizing cost simply drives through the huge bump and eventually gets flipped~(colored in \textit{red}). It demonstrates that the traversability regression component is capable of predicting a precise level of vehicle-terrain interaction, allowing for safe and effective navigation in unstructured environments.

\subsubsection{Navigational Policies}
Lastly, we further show that our framework is capable of employing different navigational objectives. The stabilizing cost is modified as the sum of the expected impacts on wheels. This would prioritize trajectories that minimize wheel impacts and attempt to avoid bumpy regions as much as possible. As illustrated in the \textit{Scenario~$\#4$}, the compact car with the modified stabilizing cost~(colored in \textit{yellow-green}) avoids all unstructured components and follows a tight path between bumps and trees at the expense of a speed reduction. It shows that our framework is not constrained to a certain navigational strategy and is scalable to various cost designs utilizing the predicted interaction.
\section{Experiments on Real-world Data}
\begin{figure}[t]
\centering
\includegraphics[width=0.99\linewidth]{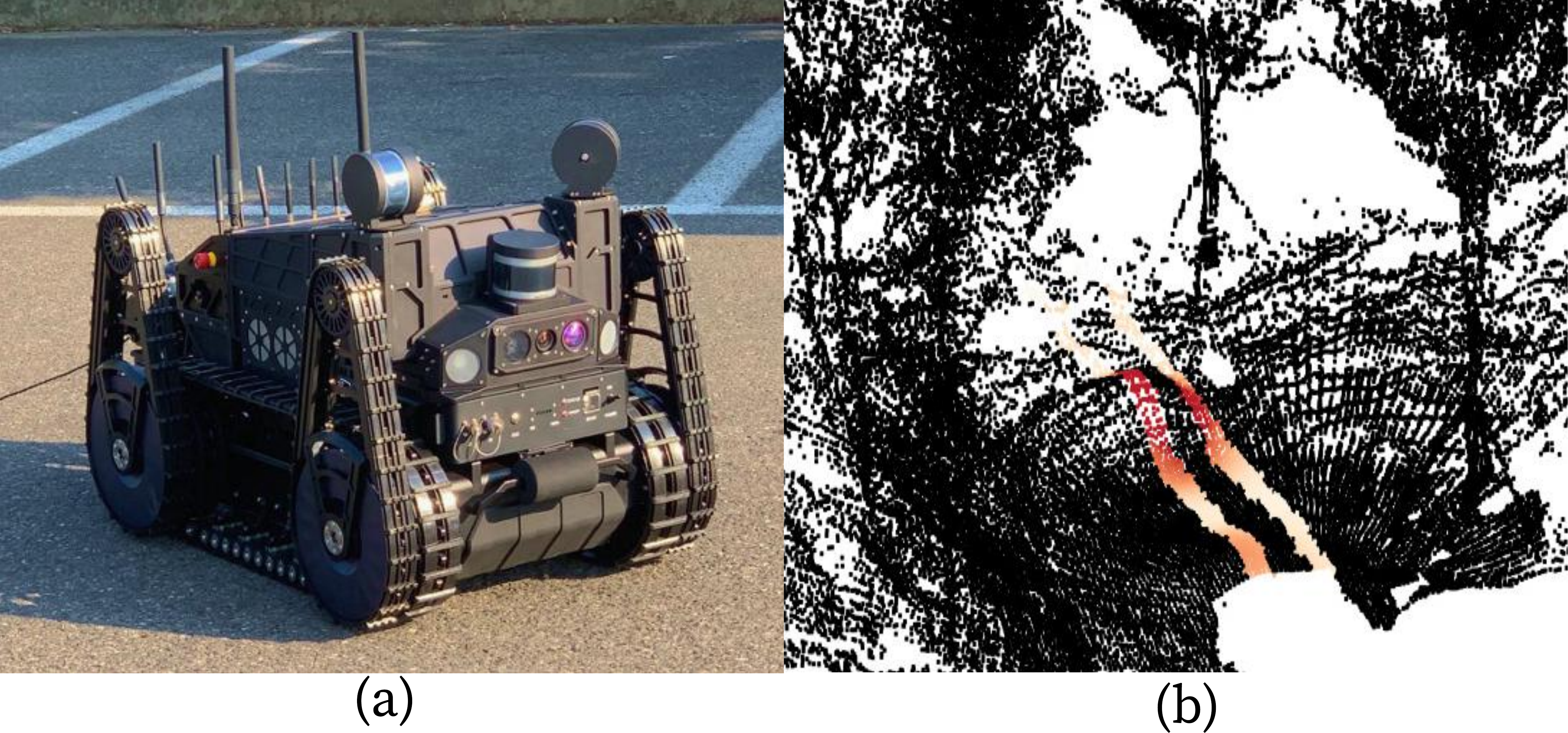}
\caption{(a) Our caterpillar-type mobility platform. (b) Mountain trail driving data. Ground-truth traversability is projected on contact points.}
\label{fig:ATE}
\vspace*{-0.2in}
\end{figure}

We also demonstrate the efficacy of our framework using real-world driving data with our mobility platform in unstructured environments and a public off-road driving dataset, RELLIS-3D~\cite{jiang2021rellis}. Specifically, we quantitatively evaluate the validity of our PU learning method for the binary classification of traversable regions in this section. We compare the following one-class classification methods:

% Also, we both quantitatively and qualitatively examine the validity of our PU learning method for the binary classification of traversable regions.

\begin{enumerate}
    \item \textit{SVDD}~\cite{ruff2018deep}, a deep-learning based anomaly detection method with the objective of $\mathcal{L}^{\text{SVDD}}$,
    \item \textit{Soft SVDD}~\cite{ruff2018deep}, which also finds a radius of a hypersphere ${R}$ that encloses the majority of the data,
    \item \textit{nn Risk}~\cite{kiryo2017positive}, a risk-estimator based PU learning algorithm with class prior set to $0.5$,
    \item \textit{Ours}, with the objective of \Eref{eq:cost}.
\end{enumerate}

\subsection{Datasets}
\subsubsection{Mountain Trail Driving Data}
We gather driving data on mountainous terrain, where not only drivable regions are not clearly specified, but also a variety of uneven terrains and hazards such as trees, rocks, bushes, and bumpy regions exist~\cite{bae2022uncertainty}. For safe driving in such environments, avoiding the regions with high uncertainty and specifying the most traversable trajectory are essential. Using our caterpillar-type vehicle, as shown in \Fref{fig:ATE}, point clouds are collected with one $32$ layer and two $16$ layer LiDARs, and the points are then fused to densely represent the environment. For ground-truth traversability, the magnitude of z-acceleration from IMU is projected to contact points. The dataset is divided into two parts: $80\%$ for training and $20\%$ for evaluation. For quantitative evaluation of the binary classification methods, only obvious traversable and non-traversable points are manually labeled as positive and negative. The remaining points are not incorporated in the evaluation.

\subsubsection{RELLIS-3D}
We also provide experimental results using a publicly available off-road dataset, RELLIS-3D~\cite{jiang2021rellis}, which contains OS1-$64$ LiDAR points with fine-grained class annotations. From the raw data, trajectories and poses are recovered using SLAM, and traversability data are generated by labeling contact points as positive. Only binary classification is trained without traversability regression. The first three out of four sequences are utilized for training, and the last sequence's data are used for evaluation with its fine-grained annotation. While the annotation does not specify which points are traversable, we define the \textit{grass} and \textit{mud} classes as positive, and the \textit{tree, vehicle, object, person, fence, barrier}, and \textit{bush} as negative.

\subsection{Quantitative Evaluation}
\begin{table}[t]
\renewcommand\arraystretch{1.2}
\caption{Quantitative results on the real-world driving data.}
\footnotesize
\centering
\newcolumntype{Y}{>{\centering\arraybackslash}m{1.8cm}}
\newcolumntype{Q}{>{\centering\arraybackslash}m{2.0cm}}
\newcolumntype{A}{>{\centering\arraybackslash}m{0.9cm}}
\newcolumntype{Z}{>{\centering\arraybackslash}m{2.0cm}}
\begin{tabular}{Z|A|Y Q}
\hline
\multirow{2}{*}{Method}& \multirow{2}{*}{Metric} & \multicolumn{2}{c}{Dataset} \\ \cline{3-4} 
                            &   & Mountain Trail & RELLIS-3D~\cite{jiang2021rellis}  \\ \hline
\textit{SVDD}~\cite{ruff2018deep}      &  \multirow{4}{*}{TPR} & 0.5000 & 0.5000 \\
\textit{Soft SVDD}~\cite{ruff2018deep} &    & 0.5000  & 0.5000 \\
\textit{nn Risk}~\cite{kiryo2017positive}   &    & 0.8540  & 0.7040 \\
\textit{Ours}      &    & \textbf{0.9426} & \textbf{0.7852} \\ \hline
\textit{SVDD}~\cite{ruff2018deep}      &  \multirow{4}{*}{AUROC}  & 0.7482          & 0.5818 \\
\textit{Soft SVDD}~\cite{ruff2018deep} &    & 0.8396  & 0.6595\\
\textit{nn Risk}~\cite{kiryo2017positive}   &    & 0.9140  & 0.7589 \\
\textit{Ours}      &    & \textbf{0.9706} & \textbf{0.7895} \\ \hline
\end{tabular}
\label{tab:Real_data_result}
\vspace*{-0.1in}
\end{table}

\begin{figure*}[t]
\centering
\includegraphics[width=0.99\linewidth]{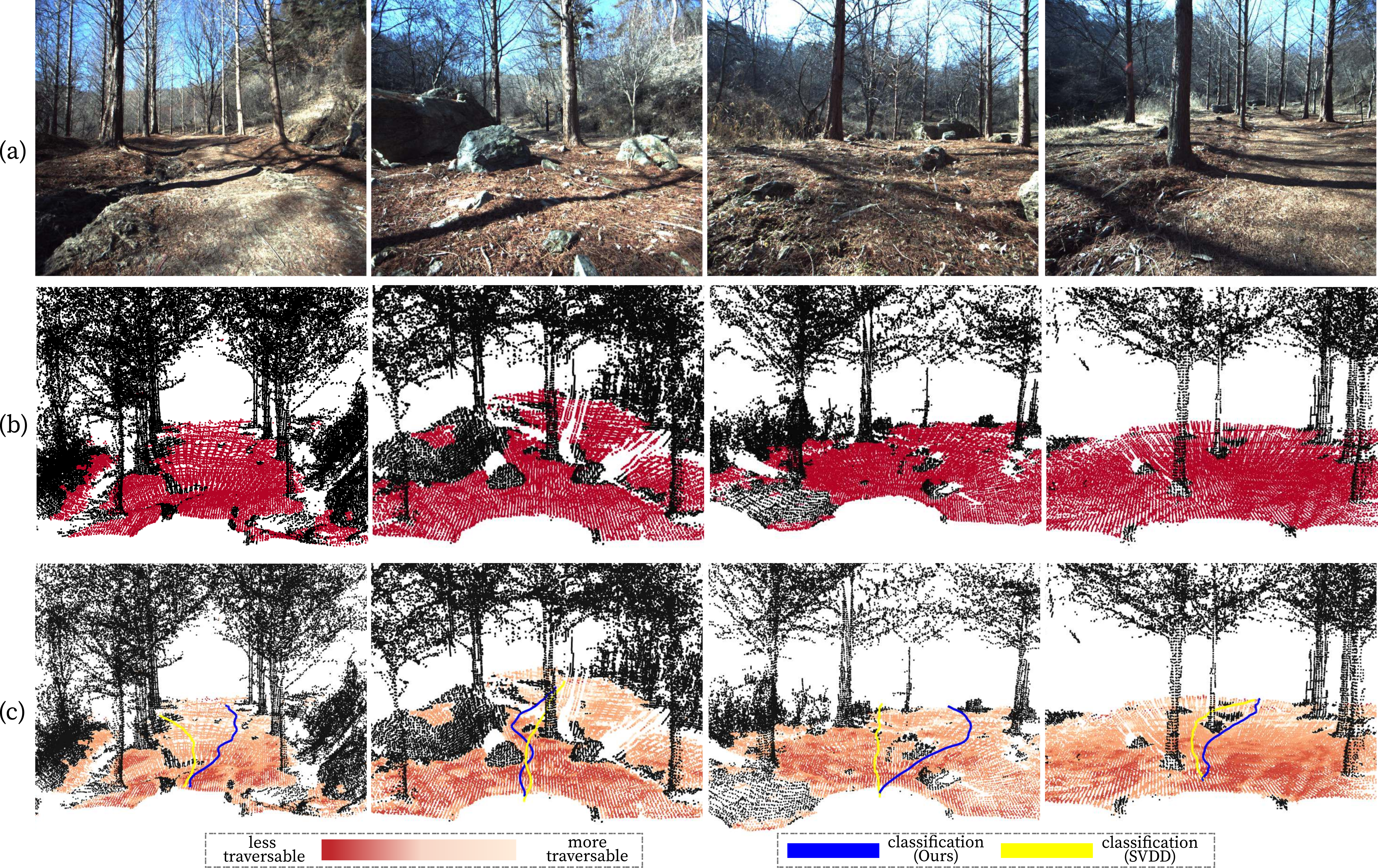}
\caption{Qualitative results for mountain trail driving data. (a) RGB images of each scene. (b) Binary classification results on 3D point clouds. The color codes are displayed in \Fref{fig:RELLIS3D}. (c) Results of traversability regression. Non-traversable regions estimated by the binary classification is masked with black. Colored lines show trajectories generated by SMPPI controller\cite{SMPPI}. Trajectories of our method successfully avoid hazardous regions.}
\label{fig:DTRAIL}
% \vspace*{-0.2in}
\end{figure*}

Area Under the Receiver Operating Characteristic~(AUROC) is used to quantitatively evaluate the quality of binary classification regardless of the threshold. AUROC quantifies the probability that a positive sample has a higher normal score than a negative sample. In addition, the mean of each class's true positive rate~(TPR) is calculated with a simple threshold $0.5$. The results are shown in \Tref{tab:Real_data_result}. 

In terms of both TPR and AUROC, our method shows a significant margin compared to other one-class classification methods for both datasets. It verifies that our method can learn more discriminative features and achieve a greater classification accuracy with a simple threshold. The TPR of anomaly detection methods is $0.5$, indicating the hypersphere collapse solution that the majority of features are trivially mapped around the positive center. Positive and negative data are differentiated marginally, and the threshold can not be determined adequately. Even though several recommendations to minimize the collapse, such as arbitrary center and no bias terms~\cite{ruff2018deep}, are followed, the point cloud encoder has not been shown to be effective in preventing the collapse. Notably, the performance of PU learning methods is better than that of anomaly detection methods, suggesting that the utilization of unlabeled data is crucial for preventing the hypersphere collapse and resolving the uncertainty problem.

\begin{figure}[t]
\centering
\includegraphics[width=0.99\linewidth]{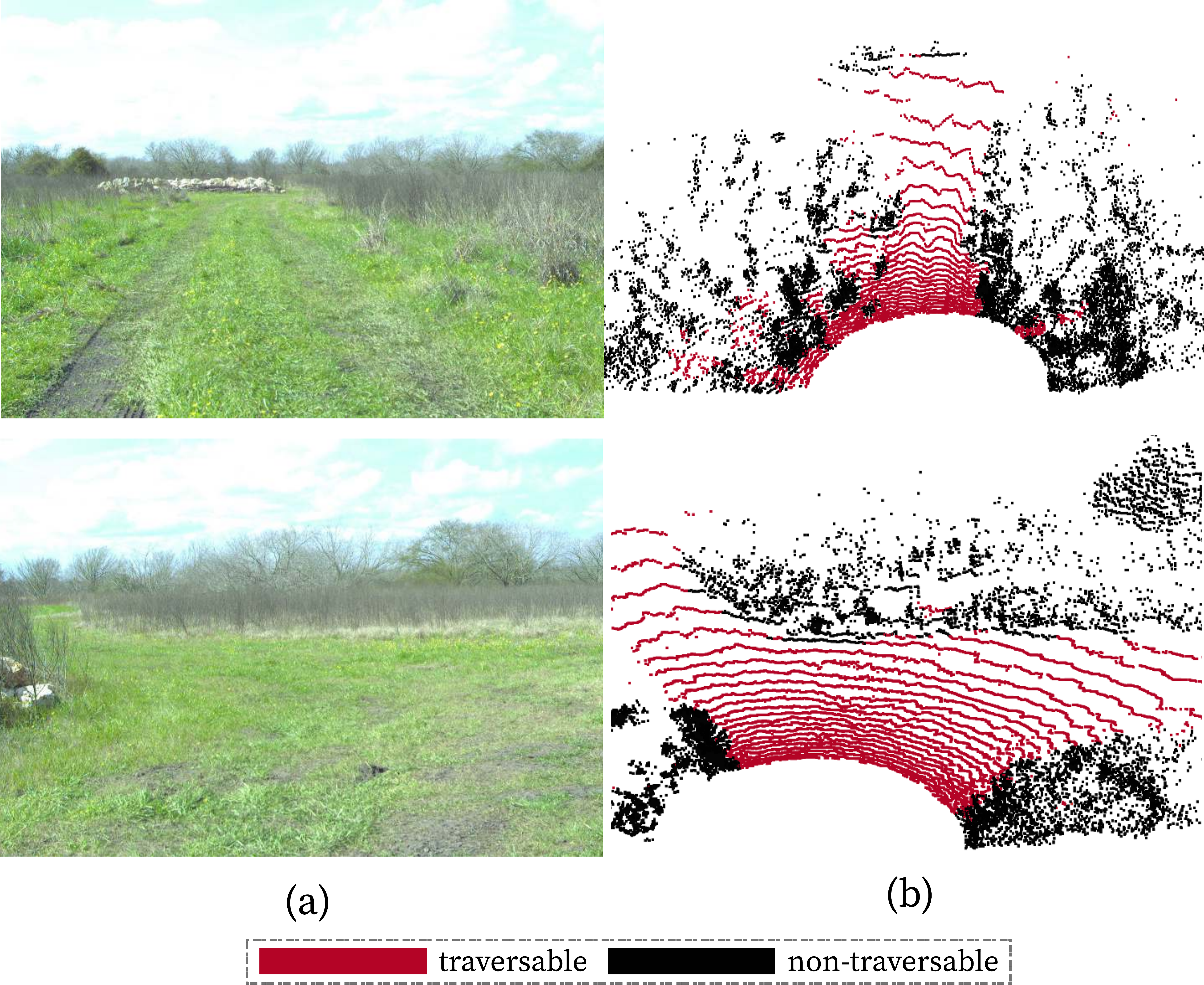}
\caption{Qualitative results for RELLIS-3D. (a) Front camera images, (b) Binary classification results.}
\label{fig:RELLIS3D}
\vspace*{-0.25in}
\end{figure}

\subsection{Qualitative Evaluation}

\Fref{fig:DTRAIL} illustrates the traversability estimation results on mountain trail driving data, and \Fref{fig:RELLIS3D} shows the binary classification results on the RELLIS-3D dataset. Rocks, wood struts, and vegetation are correctly characterized as non-traversable terrains, as they are significantly distinct from actually driven terrains.

Using the estimation results on the mountain trail driving data, trajectories are produced by the model predictive controller~\cite{SMPPI}. The cost function is identical to \Eref{eq:cost}, with the stabilizing cost simply being the estimated traversability regression value. The trajectory is also derived from the estimation result of the model trained with the anomaly detection method. As depicted in \Fref{fig:DTRAIL}, the trajectory can lead to hazardous regions such as wood struts, bumps, rocks, and gravel roads with the suboptimal result of the binary classification, whereas our method successfully avoids such non-traversable regions.

\section{Conclusions}
This paper presents a scalable framework for learning self-supervised traversability, which is an essential component for successful navigation in unstructured environments. Using the 3D point clouds, our framework learns to predict the proprioceptive experience of a vehicle when traversing the points. In addition, we address the problem of estimations being overconfident for unknown terrain. Using the PU learning method, terrains with high uncertainty are specified so that traversing such regions can be avoided. Extensive experiments demonstrated that the proposed learning framework could learn vehicle-specific traversability of various types of vehicles, which leads to successful navigation in unstructured environments. In addition, our framework does not require any human intervention and is, therefore, scalable to various vehicles, sensors, and navigational policies.

In future work, we aim to extend our framework to additional modalities, to capture terrain characteristics from visual data more effectively. In addition, while our framework requires a full training for each vehicle, we are investigating a learning method that can quickly adapt to vehicle-specific traversability using a small amount of driving data in a self-supervised fashion in order to improve scalability. Furthermore, we want to investigate which proprioceptive information can precisely model the terrain-vehicle interaction for successful navigation in unstructured environments.

% In future work, we aim to apply the self-supervised traversability learning framework to other modalities, such as images. In addition, while our framework requires a full training step for each vehicle, we are investigating a learning method that can quickly adapt to vehicle-specific traversability using a small amount of driving data in a self-supervised fashion in order to improve scalability. Furthermore, we want to investigate which proprioceptive information can precisely model the terrain-vehicle interaction for successful navigation in unstructured environments.

% \section*{Acknowledgment}
% This work was supported by the Agency For Defense Development, funded by the Korean Government in $2022$.

\addtolength{\textheight}{0cm}   % This command serves to balance the column lengths

\bibliographystyle{IEEEtran}
% \typeout{}
\bibliography{mybib }

\end{document}